\documentclass{article}

\usepackage{microtype}
\usepackage{graphicx}
\usepackage{subcaption}
\usepackage{booktabs} 

\usepackage{multirow}

\usepackage{hyperref}

\usepackage[preprint]{icml2026}

\usepackage{amsmath}
\usepackage{amssymb}
\usepackage{mathtools}
\usepackage{amsthm}

\usepackage[capitalize,noabbrev]{cleveref}

\theoremstyle{plain}

\theoremstyle{definition}

\theoremstyle{remark}

\usepackage[textsize=tiny]{todonotes}

\icmltitlerunning{
SoftMoE: Soft Differentiable Routing for Mixture-of-Experts in LLMs
}

\DeclareMathOperator{\TopK}{Top_k}
\DeclareMathOperator{\SoftTopK}{SoftTopK}
\DeclareMathOperator{\softmax}{softmax}

\usepackage{bm}

\begin{document}

\twocolumn[
  \icmltitle{
  SoftMoE: Soft Differentiable Routing for Mixture-of-Experts in LLMs
  }



  \icmlsetsymbol{equal}{*}
  
  \begin{icmlauthorlist}
    \icmlauthor{Mikołaj Zasada}{AGH}
    \icmlauthor{Łukasz Struski}{UJ}
    \icmlauthor{Jacek Tabor}{UJ,UW}
    \icmlauthor{Marcin Kurdziel}{AGH}
  \end{icmlauthorlist}

  \icmlaffiliation{AGH}{AGH University of Krakow, Poland}
  \icmlaffiliation{UJ}{Faculty of Mathematics and Computer Science, Jagiellonian University, Poland}
  \icmlaffiliation{UW}{Centre for Credible Artificial Intelligence, Warsaw University of Technology}

  \icmlcorrespondingauthor{Mikołaj Zasada}{mzasada@agh.edu.pl}

  \icmlkeywords{SoftMoE, Mixture-of-Experts, Differentiable Routing, Expert Allocation, LLM}

  \vskip 0.3in
]



\printAffiliationsAndNotice{}  

\begin{abstract}
    Sparse Mixture-of-Experts (MoE) architectures enable scaling LLM parameters under a fixed inference budget by
    activating only a small subset of experts via top-$k$ routing. While this preserves causality and suits autoregressive
    language models, the discrete top-$k$ operator is not differentiable, forcing a fixed number of active experts per
    input and resulting in inefficient use of computation. We propose SoftMoE, which replaces discrete routing with a
    truncated soft top-$k$ LapSum relaxation, allowing gradient-based optimization of expert routing. We further
    parameterize the mean number of active experts per layer and impose a global budget constraint, enabling the model
    to learn how to allocate expert capacity across layers. SoftMoE remains fully compatible with autoregressive
    modeling and achieves performance comparable to or better than sparse MoE on language modeling and downstream tasks,
    while activating significantly fewer experts. Notably, the learned allocation is highly non-uniform, with later
    layers activating more experts.
    The source code is publicly available\textsuperscript{$\dagger$}.%
    \begingroup
        \renewcommand{\thefootnote}{$\dagger$}
        \footnotetext{\url{https://github.com/dlcuda/SoftMoE}}
    \endgroup
\end{abstract}

\begin{figure}[h!]
    \centering
    \includegraphics[width=1.\columnwidth]{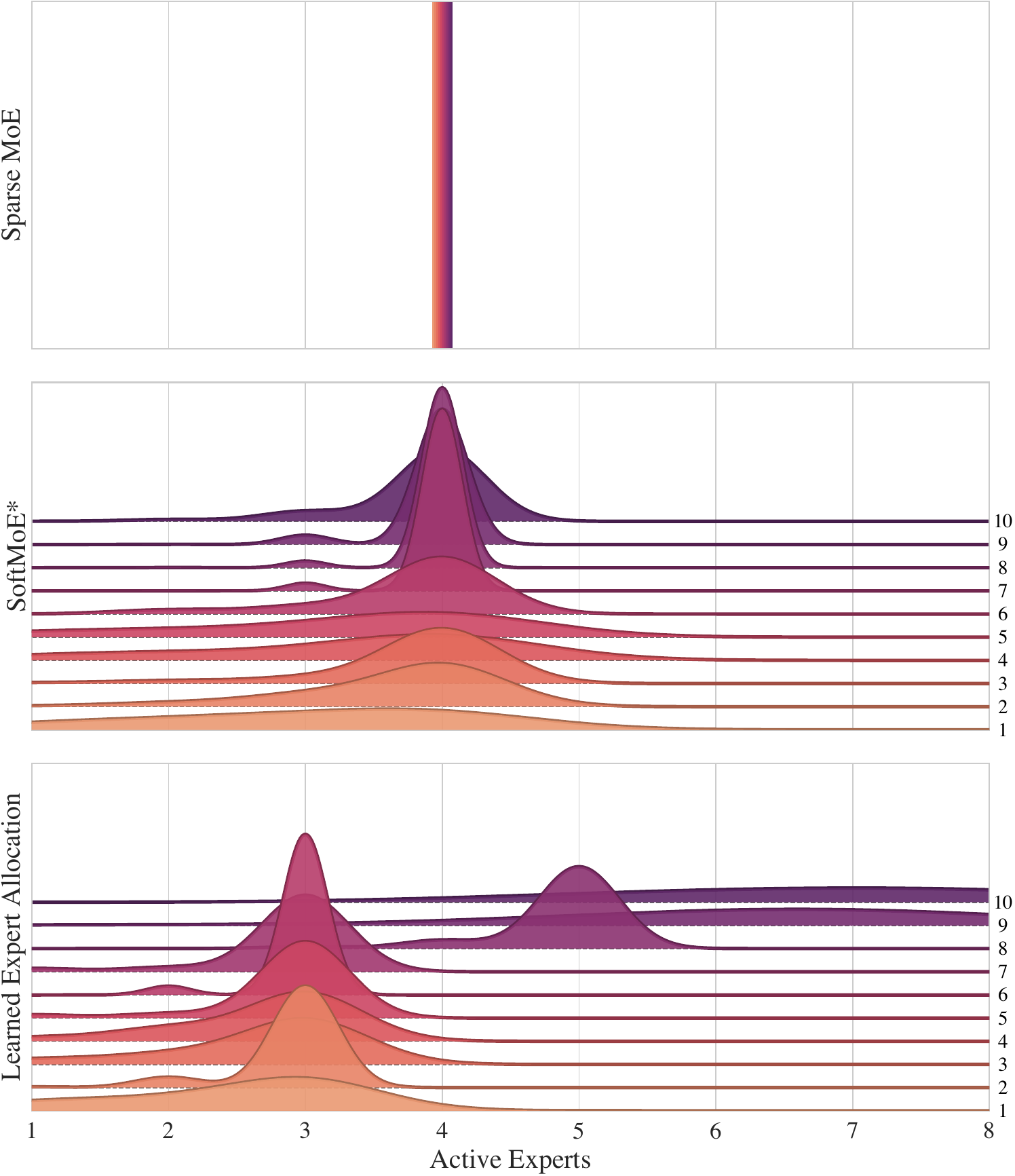}
    \caption{Learned expert allocation reveals non-uniform capacity needs across layers of a MoE model. Each color shows
             the distribution of activated experts in one layer of 10-layer model. Top: standard sparse MoE fixes the
             number of activated experts a priori. Middle: under soft top-$k$ routing, the number of activated experts
             depends on the input token. Bottom: learned allocation redistributes active experts budget: final layers
             activate more experts at the expense of initial and middle layers.}
    \label{fig:active_expert_distribution}
\end{figure}

\section{Introduction}

The rapid progress of large language models has been driven primarily by scaling model capacity, which has been
repeatedly shown to yield predictable improvements in performance~\citep{kaplan2020scaling, DBLP:journals/corr/abs-2203-15556}. While
dense transformer architectures scale effectively, their inference cost grows linearly with the number of parameters,
quickly becoming prohibitive. Mixture-of-Experts (MoE) architectures~\citep{DBLP:conf/iclr/ShazeerMMDLHD17} address
this limitation by enabling conditional computation: only a small subset of parameters is activated for each input
token, allowing models to scale to hundreds of billions of parameters while keeping the per-token computation nearly
constant.

In modern large-scale MoE models, sparse routing based on hard top-$k$ selection has emerged as the dominant
paradigm~\citep{fedus2022switch, DBLP:conf/iclr/LepikhinLXCFHKS21}. By activating only the $k$
highest-scoring experts per token, sparse MoE preserves autoregressive causality and achieves strong empirical
performance. Yet the discrete nature of the top-$k$ operator introduces fundamental limitations. Since hard top-$k$
routing is non-differentiable, gradients cannot flow through the selection process, forcing the number of active experts
to be fixed a priori. As a result, expert capacity cannot be adaptively allocated across layers or tokens, often leading
to inefficient use of computation. Moreover, these models often exhibit unstable training
dynamics~\citep{fedus2022switch}.

To address these limitations, several alternative routing strategies depart from the simple top-$k$ selection. For
example,~\citet{puigcerver2024soft} replace discrete expert selection with differentiable token mixing,
\citet{DBLP:conf/nips/ZhouLLDHZDCLL22} improve capacity allocation by allowing experts to choose tokens, and
\citet{DBLP:conf/icml/LewisBDGZ21} formulate routing as a linear assignment problem. While these approaches
alleviate some of the limitations of top-$k$ routing, they often break compatibility with autoregressive modelling,
decouple inference-time routing from the training procedure, or skew the inference-time expert load. This raises a
question: can we design a routing mechanism that adaptively allocates expert capacity across layers and tokens,
preserves causality and is computationally efficient?

In this work, we propose \emph{SoftMoE}, a Mixture-of-Experts architecture built on a differentiable relaxation
of the top-$k$ operator. Our approach replaces hard expert selection with a truncated soft top-$k$ mechanism based on the
LapSum relaxation~\citep{DBLP:conf/icml/StruskiBPT25}, enabling gradient-based optimization of expert routing.
Crucially, unlike prior soft-routing methods, SoftMoE processes each token independently and remains fully compatible
with autoregressive language modeling. In particular, SoftMoE does not require token mixing or expert-side token
selection.

Our soft routing strategy also enables learning expert allocation across layers. To this end, SoftMoE explicitly
parameterizes the mean number of active experts in each layer and impose a global constraint on the number of experts
active across the network. This mechanism allows the model to learn how to redistribute computational capacity between
layers under a fixed total inference cost. Specifically, layers that require more capacity can increase their expert
allocation at the expense of layers that require fewer active experts. Such adaptive capacity redistribution is
impossible under standard top-$k$ routing, which fixes the number of active experts a priori.

We evaluate SoftMoE on language modeling using C4 and OpenWebText, as well as downstream benchmarks. Our experiments
demonstrate that SoftMoE consistently matches or outperforms standard sparse MoE while activating significantly
fewer experts on average. Moreover, we observe that the learned capacity allocation is highly structured, with later
layers consistently utilizing more experts. This provides new empirical insights into how large language models allocate
conditional computation.

In summary, our contributions are threefold:
\begin{itemize}
\item We introduce a differentiable soft top-$k$ routing mechanism suitable for large-scale sparse MoE models.
\item We propose a learnable, globally constrained expert budget that enables adaptive allocation of experts across
  layers.
\item We demonstrate improved efficiency and competitive or superior performance on language modeling and downstream
  tasks, while maintaining autoregressive compatibility.
\end{itemize}

\section{Related Work}

Recent years have seen a gradual shift from monolithic transformer architectures toward Mixture-of-Experts (MoE) models and
conditional computation \citep{DBLP:journals/neco/JacobsJNH91, DBLP:journals/neco/JordanJ94}. The search for new
architectures was initially motivated by empirical scaling laws observed in dense large language
models~\citep{DBLP:journals/corr/abs-2203-15556, DBLP:conf/nips/MuennighoffRBST23}, which demonstrated strong
performance gains from increased model capacity. These laws were subsequently adapted to more compute-efficient sparse
architectures \citep{DBLP:conf/icml/LudziejewskiKAP24, DBLP:conf/icml/ClarkCGMPHDHCB022}, where dense layers are
decomposed into expert modules and only a subset is activated per token. This conditional computation paradigm enables a
substantial increase in model capacity while keeping the per-token computational cost nearly constant
\citep{DBLP:conf/iclr/ShazeerMMDLHD17, DBLP:conf/iclr/LepikhinLXCFHKS21}. Empirical studies further demonstrate that
sparse MoE models can outperform dense counterparts in machine translation and related sequence modeling tasks
\citep{DBLP:conf/iclr/ZhaoC0C24}. Concurrently, hybrid approaches that employ dense computation during training while
retaining sparse execution at inference, such as DS-MoE \citep{DBLP:journals/corr/abs-2404-05567}, attempt to balance
optimization stability with inference efficiency.

Despite these advantages, the effectiveness of MoE architectures is critically determined by the routing mechanism.
Suboptimal token-to-expert assignment can result in load imbalance, degraded utilization of experts, and unstable
training dynamics, which has motivated extensive work on routing simplification and stabilization strategies
\citep{fedus2022switch, zoph2022st, DBLP:conf/icml/LewisBDGZ21}. These challenges become more
pronounced at scale, where routing design choices have been shown to significantly affect convergence and performance
\citep{riquelme2021scaling}.

Historically, expert selection in MoE models has relied on hard top-$k$ routing, where each token is dispatched to the
$k$ highest-scoring experts \citep{DBLP:conf/iclr/ShazeerMMDLHD17}. While simple and effective, this discrete
selection introduces non-differentiability, motivating a range of alternative formulations. Prior work explored
expert-choice routing \citep{DBLP:conf/nips/ZhouLLDHZDCLL22}, deterministic routing rules
\citep{DBLP:conf/nips/RollerSSW21}, global optimization approaches based on linear assignment
\citep{DBLP:conf/icml/LewisBDGZ21} or optimal transport \citep{DBLP:conf/iclr/LiuPB23}, as well as reinforcement
learning–based routers \citep{DBLP:journals/corr/BengioBPP15}. However, these methods continue to rely on explicit
discretization or indirect optimization and do not fully address the mismatch between training objectives and
inference-time top-$k$ selection.

This mismatch has long been recognized in the broader context of top-$k$ learning. For example, direct top-$k$
error optimization was investigated as an alternative to top-1 optimization in
SVMs~\citep{lapin2016lossfunctionstopkerror}, with the central idea of aligning the training objective with
inference-time selection by constructing differentiable loss functions that admit gradient-based optimization. This
paradigm is relevant for MoE architectures, where routing decisions are inherently based on sparse top-$k$
selection. Subsequent work introduced smooth approximations of the top-$k$ operator that enable direct gradient
optimization and often improve accuracy \citep{berrada2018smoothlossfunctionsdeep}. However, such approaches may incur
computational overhead or trade accuracy for efficiency \citep{garcin2022stochastic}. Optimal transport–based
formulations offer smooth gradients via entropic regularisation~\citep{cuturi2019differentiable, xie2020differentiable,
masud2023multivariatesoftrank}, but their computational complexity may limit applicability to the large expert pools
in modern MoE models.

Beyond MoE, top-$k$ learning has proven effective in sparse representation learning, sparse attention in computer
vision, and recommender systems \citep{zhao2019explicitsparsetransformer, chen2019topkoffpolicy, hoefler2021sparsity,
chen2023learningsparsetransformer}, which share similarities with conditional computation. From a theoretical
standpoint, top-$k$ optimization has also been linked to statistical learning theory, with recent results providing loss
functions for optimizing cardinality of the predicted set~\citep{cortes2024cardinality, mao2024topkclassification}.

Motivated by limitations of hard top-$k$ expert selection, recent work has increasingly focused on soft routing mechanisms that relax discrete
selection into continuous, fully differentiable operations. \citet{puigcerver2024soft}, for instance, propose to
replace hard assignments with weighted combinations of tokens. Such mixing of tokens at different positions is, however,
incompatible with autoregressive modelling. \citet{DBLP:conf/nips/AntoniakKPKLCKO24} relies on batch inference and
mixes tokens from different inputs. While this enables stable end-to-end optimization, mixing tokens from different
inputs may potentially open the model to subtle cross-input attacks and information leaks. In this context,
\citet{DBLP:conf/icml/StruskiBPT25} introduced a novel family of differentiable order-statistics operators, including
soft ranking, soft top-$k$ selection, and soft permutations. Their method leverages a closed-form inverse of the LapSum
function -- defined as a sum of Laplace distributions -- allowing low-memory, low-complexity selection of the highest
activations. This makes the approach particularly well suited for large-scale MoE routing, where scalability and
efficiency are paramount.

\section{Differentiable Top-$\bm{k}$ Selection for Mixture of Experts}
\label{sec:lapsum_section}

We focus on the construction of a \emph{soft, fully differentiable} top-$k$ selection operator tailored to MoE
architectures, following the framework introduced in~\citet{DBLP:conf/icml/StruskiBPT25}. In MoE models, top-$k$
selection plays a central role in expert routing, as it determines which subset of experts is activated for a given
input. Consequently, the selection mechanism must both preserve sparsity and support efficient gradient-based
optimization.

Let $\mathbf{r} = (r_i)_{i=1}^n \in \mathbb{R}^n$ denote the routing scores produced by a gating network over $n$
experts. The hard $\TopK$ operator maps $\mathbf{r}$ to a binary routing vector:
\[
  \mathbf{p} = (p_i)_{i=1}^n \in \{0,1\}^n,
\]
where $p_i = 1$ if expert $i$ belongs to the set of the $k$ highest-scoring experts and $p_i = 0$ otherwise. This
operation enforces sparse expert activation and satisfies the cardinality constraint:
\[
\sum_{i=1}^n p_i = k.
\]
However, due to its discrete nature, hard $\TopK$ routing is non-differentiable and typically requires surrogate
gradients or heuristic approximations during training.

To overcome this limitation, we employ the differentiable relaxation of the $\TopK$
operator~\citep{DBLP:conf/icml/StruskiBPT25} that replaces the binary routing vector with a continuous assignment: \[
  \tilde{\mathbf{p}} = (\tilde{p}_i)_{i=1}^n \in [0,1]^n,
\]
interpreted as soft expert selection weights. The relaxed operator preserves the expected sparsity pattern through the
constraint:
\[
\sum_{i=1}^n \tilde{p}_i = k,
\]
for a (possibly non-integer) selection parameter $k$. 

\begin{figure}[thb]
    \centering
    \includegraphics[width=1\linewidth]{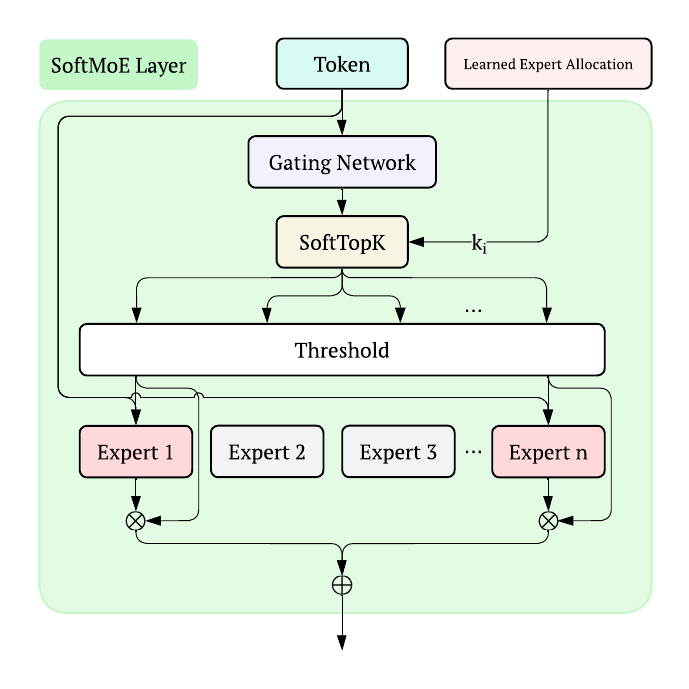}
    \caption{Architecture of a SoftMoE layer. The dense MLP layers in each transformer block are replaced by smaller
             expert MLPs with soft routing (light green box). The router computes scores over all experts, which are
             transformed by the SoftTopK operator into soft selection weights $p_i \in [0,1]$. Weights below threshold
             are truncated to zero, yielding sparse but input-dependent expert activation.}
    \label{fig:flowchart_soft_moe_layer}
\end{figure}

The construction formulates top-$k$ routing as a continuous order-statistics problem and introduces the \emph{LapSum}
function, defined as a sum of Laplace cumulative distribution functions. The method identifies a differentiable
threshold $x$ by solving:
$$
\mathrm{LapSum}(x) = \sum_{i=1}^n F_{\text{Lap}}(r_i - x) = k,
$$
where $r_i$ are the input scores and $F_{\text{Lap}}$ denotes the Laplace CDF. This equation admits a unique solution $\tilde{x}$, which can be computed efficiently in closed form, without sorting or iterative optimization. As a result, the method achieves linear time and memory complexity with respect to the number of experts, while remaining fully differentiable with respect to both the routing scores $r$ and the selection parameter~$k$. We later exploit the latter property to optimize expert allocation patterns across MoE layers.

Intuitively, $\mathrm{LapSum}(x)$ acts as a smooth approximation of the counting function $\sum_i \mathbf{1}[r_i \ge x]$, and solving $\mathrm{LapSum}(x)=k$ corresponds to finding a continuous relaxation of the $k$-th order statistic (i.e., the top-$k$ threshold). The resulting soft selection weights are then given by $F_{\text{Lap}}(r_i - \tilde{x})$, which provide a differentiable approximation of the binary top-$k$ mask. This formulation can be viewed as a generalization of the softmax operator: similarly, it produces an output tensor of the same dimensionality as the input. However, unlike softmax (where the outputs form a probability distribution summing to 1), LapSum enforces that the outputs sum exactly to $k$, while each individual element remains in the range $[0, 1]$. This makes it particularly suitable for differentiable top-$k$ selection, where both sparsity and gradient flow are required.

\section{From Discrete to Soft Expert Selection}\label{sec:soft_routing}

In Section~\ref{sec:lapsum_section}, we introduced a differentiable relaxation of the top-$k$ operator that maps routing
scores $\mathbf{r} \in \mathbb{R}^n$ to soft selection weights $\mathbf{\tilde{p}} \in [0,1]^n$, such that $\sum_1^n
\tilde{p}_i = k$. We now apply this relaxed $\SoftTopK$ operator to Mixture-of-Experts routing.

Let $\mathbf{x} \in \mathbb{R}^d$ be an input token representation. In a standard sparse MoE architecture, the routing
network selects which expert to activate for $\mathbf{x}$ using a discrete gating function:
\begin{equation*} \label{eq:hard_gating}
  G(\mathbf{x}) = \TopK\bigl(\softmax(\mathbf{x} \mathbf{W}_g), k\bigr),
\end{equation*}
where $\mathbf{W}_g \in \mathbb{R}^{d \times n}$ is a learnable gating matrix and $n$ is the number of experts. The layer
output is then calculated as a weighted combination of the experts' computation:
\begin{equation*} \label{eq:moe_output}
  \mathbf{y} = \sum_{i=1}^{n} G(\mathbf{x})_i \cdot E_i(\mathbf{x}),
\end{equation*}
where $E_i: \mathbb{R}^d \to \mathbb{R}^d$ is the $i$-th expert network. Sparsity of the $\TopK$ operator guarantees
that exactly $k$ experts contribute to the output. However, because $\TopK$ is non-differentiable, training gradient
cannot flow through the expert selection process. Consequently, $\mathbf{W}_g$ is trained using the surrogate gradient
from the pre-selection $\softmax$ scores.

In SoftMoE (Figure~\ref{fig:flowchart_soft_moe_layer}), we replace the discrete $\TopK$ operator with its differentiable
LapSum relaxation:
\begin{equation*} \label{eq:soft_gating}
  G(\mathbf{x}) = \mathcal{T}\bigl[\SoftTopK(\mathbf{x} \mathbf{W}_g, k)\bigr],
\end{equation*}
where $\SoftTopK(\cdot, k)$ is the soft top-$k$ operator introduced in Section~\ref{sec:lapsum_section}. While the soft
top-$k$ operator provides differentiability, naively routing through all $n$ experts using all soft weights would result
in dense computation. Our goal, however, is to adapt the number of active experts to the input token and, subsequently,
to learn how to allocate computational budget across layers. To recover computational efficiency, we therefore truncate
the LapSum relaxation to remove low-contribution weights:
\begin{equation*} \label{eq:soft_gating2}
  \mathcal{T}(z) = z\mathbb{I}_{z > \tau},
\end{equation*}
where $\mathbb{I}$ is the indicator function. For $z > \tau$, the gradient passes through the operator unchanged, while
for $z < \tau$, the gradient is zero. At $z=\tau$ we adopt the subgradient convention $\mathcal{T}'(0) = 0$.

The truncation operator $\mathcal{T}(z)$ is equivalent to a shifted ReLU: active experts receive gradient through their routing
weights, while inactive experts receive none. At the decision boundary, expert selection is non-differentiable,
analogous to hard top-$k$ routing. That said, the model remains differentiable with respect to the selection parameter
$k$ via the LapSum inverse. This gradient pathway is unavailable when $k$ is a fixed integer, and is what enables learning
per-layer expert allocation under the global budget constraint. Moreover, because truncation applies a threshold rather
than selecting by rank, the number of active experts is input-dependent, allowing the model to adjust per-token compute.
In summary, SoftMoE is a routing mechanism that enables gradient-based optimization of per-layer expert budgets and
per-token compute adaptation, rather than a differentiable replacement of hard top-$k$ selection.

\paragraph{Learning Expert Allocation Across Layers.}

A key advantage of the soft gating function is that the model becomes differentiable with respect to the expected number
of active experts $k$. Consequently, we can \emph{learn} how many experts to activate in each layer, rather than fixing
this choice a priori. Such learned allocation of experts is impossible with standard sparse MoE architecture, which
typically activates the same number of experts in each layer. 

Consider a network with $L$ MoE layers. We parametrize expert allocation across layers with a vector $\mathbf{k} =
\bigl(k_l\bigr)_{l=1}^L$, where $k_l$ denotes the mean number of active experts in layer $l$. Obviously, we
require at least one active expert in each layer, i.e. $k_l \geq 1,\ l \in \bigl\{1, \ldots L\bigr\}$. Additionally, we
need a constraint that prevents the network from learning to use all available experts, which would be equivalent to a
dense computation. An intuitive way to achieve this is to introduce a global budget for active experts, and then let the
optimization process allocate that budget across layers. We therefore optimize expert allocation under the following
constraints:

\begin{equation} \label{eq:budget_constraint}
  \begin{aligned}
      & k_l \geq 1, \quad l \in \{1, \ldots, L\} \\
      & \sum_{l=1}^{L} k_l = K
  \end{aligned},
\end{equation}

where $K$ is the global budget for active experts across all layers. This formulation ensures that the total
computational cost remains controlled. With a budget of $K = 2L$, for instance, the average compute matches a sparse MoE
model with $k=2$ active experts per layer. That said, we expect learned allocation to match performance of a sparse MoE
while using a tighter budget on active experts.

The constraints in Eq.~\ref{eq:budget_constraint} can be easily reformulated in an unconstrained domain. To this end, we
introduce auxiliary parameters $\boldsymbol{\eta} = \bigl(\eta_l\bigr)_{l=1}^L \in \mathbb{R}^L$ and reparametrize
expert allocation as:
\begin{equation*} \label{eq:budget_constraint_rep}
  \boldsymbol{\pi} = \softmax(\boldsymbol{\eta}), \quad  k_l = \pi_l \cdot (K - L) + 1.
\end{equation*}
Because $\softmax$ ensures $\sum_l \pi_l = 1$, the affine transform maps the probability simplex to the feasible region
where $\sum_l k_l = K$ and each $k_l \geq 1$. The reparametrization is end-to-end differentiable, enabling optimization
of unconstrained allocation parameters $\boldsymbol{\eta}$ jointly with other model parameters, using standard gradient
descent.

The budget constraint in Eq.~\ref{eq:budget_constraint} introduces competition between layers: increasing the expert
allocation in one layer requires matching compensation in other layers. Intuitively, this forces the model to
redistribute computation to layers that benefit most from additional expert capacity. As we demonstrate in
Section~\ref{sec:experiments}, this competition leads to highly non-uniform allocations. Importantly, learned expert
allocation does not introduce any computational bottleneck: the available compute units can be utilized uniformly, by
assigning units to layers proportionally to the learned means~$k_l$.

\begin{figure}[thb]
    \centering
    \includegraphics[width=1\linewidth]{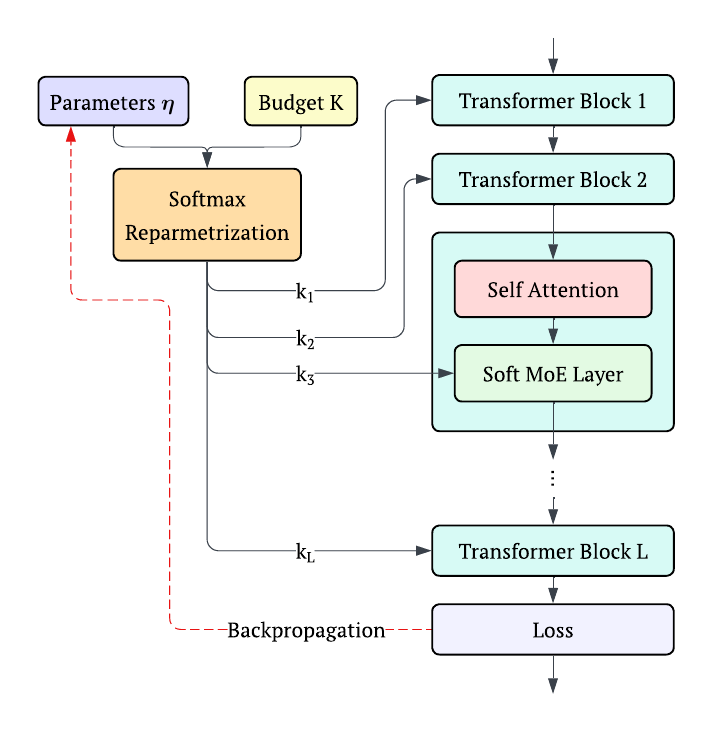}
    \caption{Transformer model with learned per-layer experts allocation and SoftMoE routing. Expert allocation is
             optimized with respect to unconstrained parameters $\boldsymbol{\eta}$. Softmax reparametrization ensures
             that each layer has at least one expert allocated ($k_l \geq 1$) and enforces global constraint on active
             experts ($\sum_l k_l = K$). Expert budget redistribution is end-to-end differentiable.}
    \label{fig:flowchart_soft_moe_transformer}
\end{figure}

Figure~\ref{fig:flowchart_soft_moe_transformer} illustrates the SoftMoE transformer architecture with learned allocation
parameters.

\paragraph{Computational Overhead.}

Operations in SoftMoE router add minimal overhead compared to sparse MoE routing. LapSum has $O(n)$ computational cost,
where $n$ is the number of experts. Memory usage is also $O(n)$ per token vs. $O(k)$ for sparse MoE. Given typical expert
counts in MoE layers ($n = 32$--$64$), this is negligible compared to the cost of multiplying by $n \times d$ gating
matrix in both routers, and does not negate benefits of reduced average number of active experts.

\begin{figure*}[thb]
    \centering
    \includegraphics[width=0.95\linewidth]{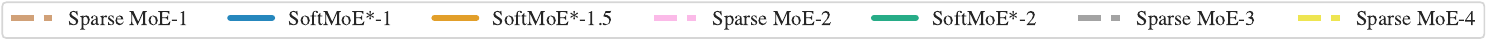}

    \begin{subfigure}{0.49\textwidth}
        \centering
        \includegraphics[width=\linewidth]{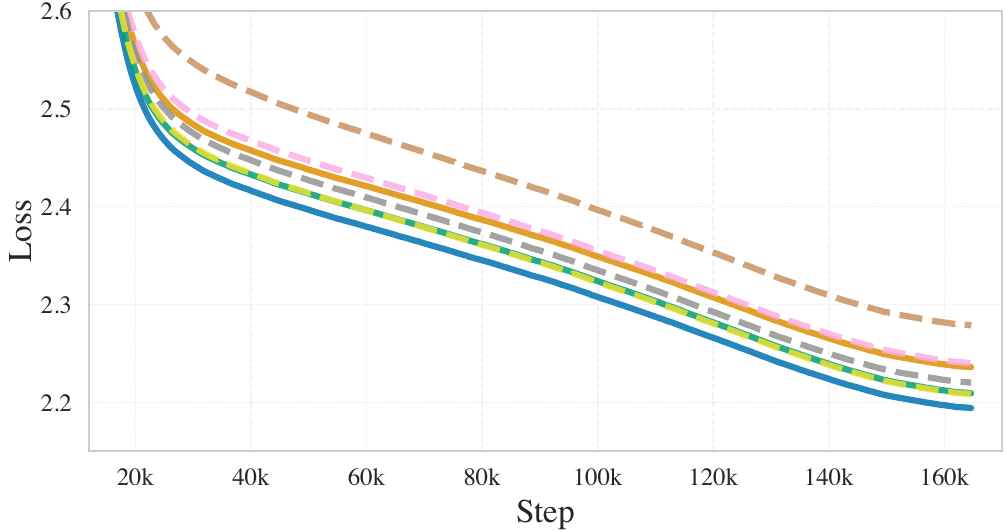}
    \end{subfigure}
    \hfill
    \begin{subfigure}{0.49\textwidth}
        \centering
        \includegraphics[width=\linewidth]{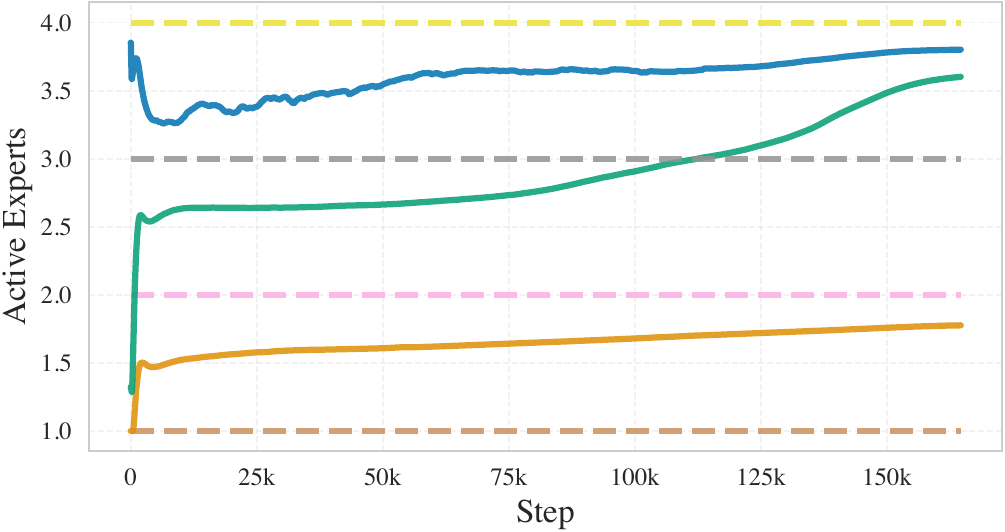}
    \end{subfigure}
    \caption{Training loss (left) and average number of active experts per token per layer (right) for Sparse MoE and
             SoftMoE* routing~(no learned expert allocation across layers). Top-$k$ selection fixes the number of
             active experts a priori. Under soft routing, active experts count depends on the distribution of soft
             weights and evolves during training.}
    \label{fig:hard_vs_soft_loss_and_ae}
\end{figure*}

\begin{figure*}[thb]
    \centering
    \includegraphics[width=0.7\linewidth]{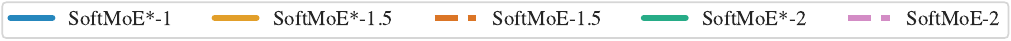}

    \begin{subfigure}{0.49\textwidth}
        \centering
        \includegraphics[width=\linewidth]{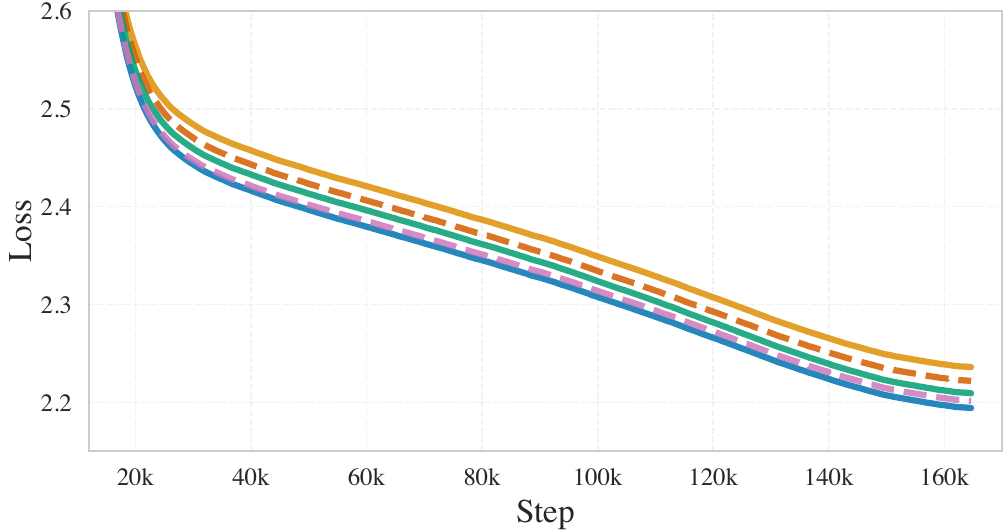}
    \end{subfigure}
    \hfill
    \begin{subfigure}{0.49\textwidth}
        \centering
        \includegraphics[width=\linewidth]{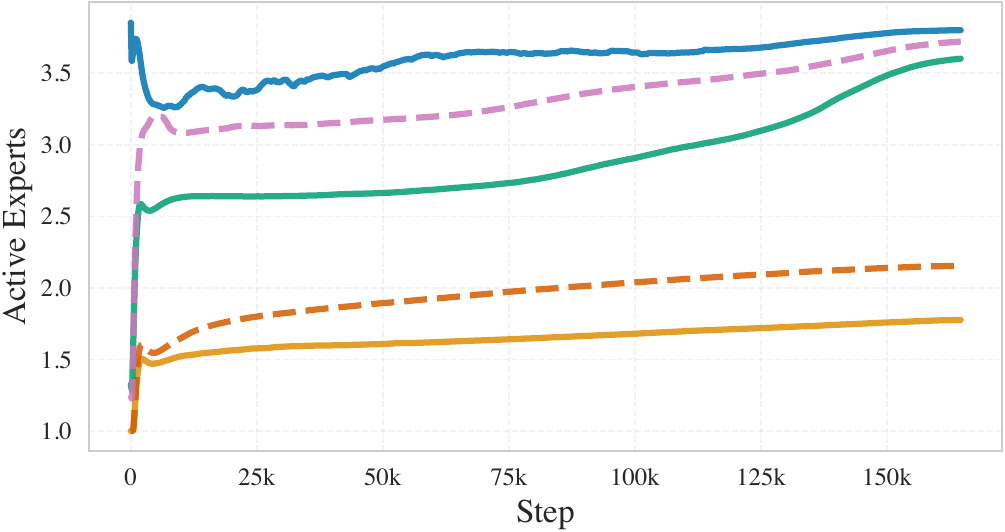}
    \end{subfigure}
    \caption{Training loss (left) and average number of active experts per token per layer (right) for soft routing
             without learning expert allocation (SoftMoE\mbox{*}) and with learned allocation (SoftMoE).}
    \label{fig:soft_vs_learning_loss_and_ae}
\end{figure*}

\section{Experiments} \label{sec:experiments}

We evaluate SoftMoE on autoregressive language modelling and downstream tasks. The experiments address two questions.
First, whether SoftMoE routing matches or exceeds sparse MoE performance while using fewer active experts. Second, what
allocation patterns emerge when the model is free to optimize expert allocation across layers under a fixed global
budget. To answer these questions we compare SoftMoE against the Switch Transformer~\citep{fedus2022switch}
architecture, the de-facto standard MoE architecture for autoregressive language models. Except for different routing
networks, all evaluated models share the same base architecture and pretraining protocol.

\paragraph{Datasets and Downstream Tasks.}

We train all models on two English text corpora of different scales: Open Web Text (OWT) \cite{Gokaslan2019OpenWeb} and Common
Crawl (C4) \cite{DBLP:journals/jmlr/RaffelSRLNMZLL20}. For each corpus, we train a BPE tokenizer with a vocabulary of
32k tokens, yielding approximately 9B tokens for OWT and 206B tokens for C4. In each case we pretrain on 98\% of the
corpus and use the remaining 2\% for validation. 

After training, we evaluate each model on three zero-shot benchmarks: PIQA~\cite{DBLP:conf/aaai/BiskZLGC20} for physical
commonsense reasoning, HellaSwag~\cite{DBLP:conf/acl/ZellersHBFC19} for sentence completion, and ARC-E
\cite{DBLP:journals/corr/abs-2102-03315} for elementary science questions. These tasks assess different reasoning
capabilities and complement language modelling loss-based evaluation.

\paragraph{Model Architecture and Training Protocol.}

All evaluated models follow a decoder-only GPT-2 architecture~\cite{radford2019language}, with dense MLP layers replaced
by, respectively, Switch Transformer-based sparse MoE or SoftMoE layers. In each case we use a network with 10
transformer blocks and 32 experts per block. Each expert contains 5M parameters, yielding a total of 1.63B trainable
parameters. We train standard sparse MoE models with $k \in \bigl\{1, 2, 3, 4\bigr\}$ active experts per layer, and
SoftMoE models with initial mean number of active experts $k \in \bigl\{1, 1.5, 2\bigr\}.$ We also train a variant of
SoftMoE models with soft top-$k$ routing, but without learned expert allocation (SoftMoE\mbox{*}).

We train on C4 for 164k steps and on OWT for 18k steps. To maintain balanced expert load in each layer, all models are
trained with auxiliary balancing loss~\citep{fedus2022switch}. Following standard practice, we also add small noise to
the pre-activation values in MoE gating functions, encouraging networks to explore different routing patterns. All
training runs were executed using Megatron-LM~\citep{DBLP:journals/corr/abs-1909-08053} framework extended with SoftMoE
routing. Complete training hyperparameters are provided in Appendix~\ref{app:train_param}.

\paragraph{Implementation Details.} \label{subsec:thresholding}

For computational efficiency, SoftMoE requires truncating low-contribution weights (Section~\ref{sec:soft_routing}). In
practice, we set the truncation threshold $\tau$ by collecting activation statistics from initial input batches and
adjusting the threshold so that initially the routing networks activate, on average, approximately $k$ experts per
layer per token. Additionally, in each layer $i \in \bigl\{1, \ldots, L\bigr\}$ we upper-bound the number of active
experts by $\lceil k_i \cdot \alpha \rceil$, where $k_i$ is the expected number of active experts and $\alpha \in
\bigl\{2, 4\bigr\}$ is an expansion factor. This mechanism prevents OOM errors on tokens with near-uniform routing
scores, while still allowing input-dependent expert selection to activate several-fold more experts than the learned
mean.

\begin{figure}[thb]
    \centering
    \includegraphics[width=1\linewidth]{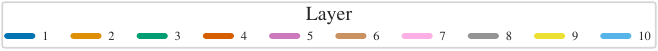}
    \includegraphics[width=1\linewidth]{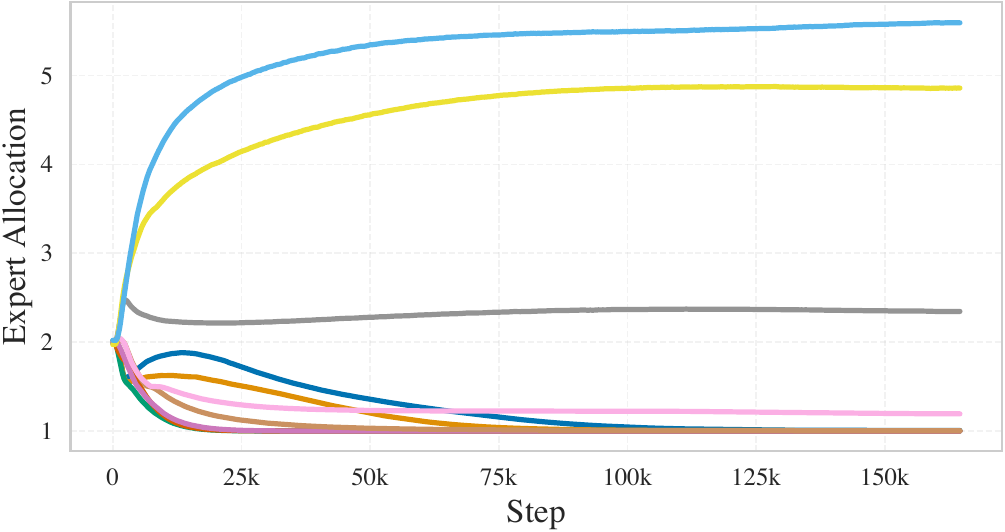}

    \caption{Evolution of the expert allocation parameters $k_l$ during training. SoftMoE architecture redistributes the active expert budget towards the final layers of the model.}
    \label{fig:k_while_learnign_k_in_trainign}

\end{figure}

\begin{table*}[thb]
    \centering
    \caption{Comparison of SoftMoE, SoftMoE with no learned expert allocation (SoftMoE$\mbox{*}$) and Sparse MoE models
             on OWT and C4 datasets. For each configuration, the table reports mean activated experts count during
             training (Train-AE) and inference (Infer-AE), mean activated parameters count (in millions) during inference (Active-Param),
             accuracy on the downstream tasks, validation loss and MoE layer configuration ($k$ and expansion factor
             $\alpha$).  Results are grouped by the budget on average number of active experts per layer per token.}
    \label{tab:owt_and_c4_results}

    \begin{tabular}{@{}c@{\;}l@{\;\,}r@{\quad}c@{\quad}l@{\;\,}l@{\;\,}c@{\;\,}c@{\;\,}c@{\;\,}c@{\;\,}c@{\;\,}c@{}}
        \toprule
         Data & \multicolumn{1}{c}{Model} & \multicolumn{1}{c}{$k$} & $\alpha$ & Train-AE & Infer-AE & Active-Param & PIQA & HELLA & ARC-E & Loss & \#Expert \\
        \midrule
        
        \multirow{9}{*}{
          \rotatebox[origin=c]{90}{Open Web Text}
        }
        
        & Sparse MoE & 1 & - & 1 & 1 & 107.74 & 61.53 & 30.56 & 33.33 & 2.84 & $= 1$ \\[0.4em]
        
        & Sparse MoE & 2 & - & 2 & 2 & 156.94 & 62.40 & 31.50 & 34.74 & 2.79 & $\leq 2$ \\
        & SoftMoE\mbox{*} & 1.5 & 2 & 1.53 & 1.73 & 143.66 & \textbf{62.62} & \textbf{32.13} & \textbf{35.10} & \textbf{2.78} & $\leq 2$ \\
        & SoftMoE & 1.5 & 2 & 1.63 & 1.96 & 154.98 & \textbf{62.57} & \textbf{32.12} & \textbf{35.45} & \textbf{2.76} & $\leq 2$ \\[0.4em]
        
        & Sparse MoE & 3 & - & 3 & 3 & 206.14 & 63.55 & 32.16 & 34.92 & 2.77 & $\leq 3$ \\[0.4em]
        
        & Sparse MoE & 4 & - & 4 & 4 & 255.34 & 63.87 & 32.36 & 36.68 & 2.75  & $\leq 4$ \\
        & SoftMoE\mbox{*} & 1 & 4 & 3.64 & 3.73 & 242.06 & \textbf{63.93} & \textbf{33.68} & 36.51 & \textbf{2.70} & $\leq 4$ \\
        & SoftMoE\mbox{*} & 2 & 2 & 2.62 & 3.12 & 212.05 & \textbf{63.82} & \textbf{32.50} & \textbf{37.21} & \textbf{2.75} & $\leq 4$ \\
        & SoftMoE & 2 & 2 & 3.18 & 3.50 & 230.74 & 63.55 & \textbf{32.74} & \textbf{38.27} & \textbf{2.74} & $\leq 4$ \\

        \midrule
        
        \multirow{9}{*}{
          \rotatebox[origin=c]{90}{C4}
        }

        & Sparse MoE & 1 & - & 1 & 1 & 107.74 & 69.75 & 43.14 & 42.5 & 2.27 & $= 1$ \\[0.4em]

        & Sparse MoE & 2 & - & 2 & 2 & 156.94 & 71.98 & 45.41 & 40.74 & 2.24 & $\leq 2$ \\
        & SoftMoE\mbox{*} & 1.5 & 2 & 1.65 & 1.81 & 147.60 & 71.06 & \textbf{45.49} & 39.15 & \textbf{2.23} & $\leq 2$ \\[0.4em]

        & Sparse MoE & 3 & - & 3 & 3 & 206.14 & 71.60 & 46.36 & 42.50 & 2.22 & $\leq 3$ \\
        & SoftMoE & 1.5 & 2 & 1.96 & 2.19 & 166.29 & 71.38 & \textbf{46.79} & \textbf{43.39} & \textbf{2.22} & $\leq 3$ \\[0.4em]

        & Sparse MoE & 4 & - & 4 & 4 & 255.34 & 71.60 & 47.23 & 43.56 & 2.20 & $\leq 4$ \\
        & SoftMoE\mbox{*} & 1 & 4 & 3.60 & 3.82 & 246.49 & \textbf{72.91} & \textbf{48.48} & 43.21 & \textbf{2.19} & $\leq 4$ \\
        & SoftMoE\mbox{*} & 2 & 2 & 2.91 & 3.62 & 236.65 & \textbf{71.60} & \textbf{48.05} & 42.86 & 2.21 & $\leq 4$ \\
        & SoftMoE & 2 & 2 & 3.35 & 3.76 & 243.54 & \textbf{72.03} & \textbf{47.48} & 42.15 & \textbf{2.20} & $\leq 4$ \\
        \bottomrule 
    \end{tabular}
\end{table*}

\subsection{Results}

\paragraph{SoftMoE\mbox{*} vs. Sparse MoE.}

We begin evaluation with comparison of sparse MoE baselines against SoftMoE routing without learning expert allocation
(SoftMoE\mbox{*} models). Table~\ref{tab:owt_and_c4_results} reports results grouped by active expert budgets. Soft
routing improves efficiency: across all configurations SoftMoE\mbox{*} consistently achieves lower language modelling loss
while activating fewer experts on average. The pattern is particularly evident with the $\leq 2$ experts per layer
per token budget, where SoftMoE\mbox{*} improves modelling loss while activating between $17\%$ (C4) and $24\%$ (OWT) fewer
experts in training. Similar pattern is evident in inference, with SoftMoE\mbox{*} consistently activating fewer experts
on average.

This efficiency gain likely stems from differentiable routing. Specifically, SoftMoE\mbox{*} can adapt the distribution
of gating weights to the token representation, rather than committing to activate exactly $k$ experts per each input.  In
particular, given a simple token the routing network can concentrate the weights on just one expert, whereas a token
that benefits from additional capacity may be routed through a more uniform weight distribution, activating many experts.
Indeed, we observe exactly this effect on empirical distributions of activated experts, especially in the initial layers
of the model (Figure~\ref{fig:active_expert_distribution}, middle plot).

Training dynamics for both architectures is shown in Figure~\ref{fig:hard_vs_soft_loss_and_ae}. Unlike sparse MoE,
which fixes active expert count a priori, SoftMoE\mbox{*} adapts the number of activated experts during training, as the
model learns which tokens benefit from more or less compute. This adaptation does not introduce any training
instabilities.

\paragraph{Learned Expert Allocation.}

Next, we focus on SoftMoE models with learned expert allocation. Figure~\ref{fig:k_while_learnign_k_in_trainign} reveals
the typical allocation pattern learned on C4. Results for other configurations are reported in
Appendix~\ref{app:additional_results}. When given freedom to optimize expert allocation across layers, SoftMoE discovers
markedly non-uniform allocation patterns. In particular, under a global constraint on active experts budget, SoftMoE
redistributes expert capacity from early and middle layers towards final layers. Concretely, in our experiments the top
3 transformer layers absorb approximately 50\% of the total expert budget. This allocation pattern arises rapidly in
initial stages of training, typically redistributing most of the budget within the first 25k training steps. The
learned pattern translates to non-uniform expert allocation in inference (Figure~\ref{fig:active_expert_distribution},
bottom plot).

The observed non-uniform allocation suggests that the model benefits from increased capacity in the final stages of token
processing. This observation aligns with prior work on information processing in transformer models. For
example~\citep{tenney2019bert} used probing tasks to characterize representations across transformer network depth, and
demonstrated that deeper layers of BERT capture high-level semantic information, while earlier layers encode simpler
syntactic patterns. Related work by~\citet{jawahar2019bert} found that early layers of BERT models encode
phrase-level information, while top layers encode semantic features. The learned expert allocation suggests similar
phenomenon in decoder-only MoE models, with layers encoding semantic information benefiting from more expert capacity,
and layers at initial stages of processing functioning effectively while activating fewer experts.

Our findings may have notable practical applications. Currently used MoE architectures usually adopt uniform
expert allocation across layers. Our results, however, suggest that depth-aware allocation may be more appropriate.

Figure~\ref{fig:soft_vs_learning_loss_and_ae} compares training dynamics with (SoftMoE) and without (SoftMoE\mbox{*})
learned expert allocation. Table~\ref{tab:owt_and_c4_results} reports language modelling loss for both variants.
Learned expert allocation typically improves language modelling performance, though with a higher, on average, number of
activated experts. This trade-off reflects the model's preference for concentrating additional capacity in later layers.

\paragraph{Downstream Evaluation.}

Downstream task performance confirms the benefits observed in language modelling loss
(Table~\ref{tab:owt_and_c4_results}). On the largest HellaSwag benchmark, soft routing consistently outperforms Sparse
MoE across all evaluated configurations on both pretraining datasets. Results for the PIQA benchmark follow a similar
pattern, with soft routing achieving better or comparable accuracy in most configurations, while activating fewer
experts. Importantly, configurations with the best pretraining loss tend to achieve strongest results on these two
benchmarks, indicating that soft routing benefits translate to task-level performance.  ARC-E exhibits higher variance
across configurations, which we attribute to its smaller evaluation set. Nonetheless, SoftMoE achieves the strongest
ARC-E performance under OWT pretraining and remains competitive under C4 pretraining, while activating fewer experts.

\section{Conclusions}

In this work, we presented SoftMoE, a Mixture-of-Experts architecture that replaces hard top-$k$ routing with a soft
top-$k$ relaxation based on the LapSum operator. By addressing the non-differentiability inherent in standard sparse
MoE routing, the proposed approach enables end-to-end gradient-based optimization of expert selection while maintaining
sparsity, computational efficiency, and compatibility with autoregressive language modeling. This directly addresses the
long-standing discrepancy between training-time optimization and inference-time routing decisions in MoE models.

A central property of SoftMoE is its ability to learn the allocation of expert capacity across layers under a fixed
global computational budget. Rather than enforcing a predefined number of active experts per layer, the model is allowed
to dynamically redistribute computation based on learned routing behavior. Empirical results indicate that this leads to
highly non-uniform expert utilization patterns, with later layers consistently receiving a larger share of the
computational budget. These observations provide additional insight into how conditional computation is exploited in
deep language models.

Experimental evaluation on language modeling and downstream reasoning benchmarks shows that SoftMoE achieves performance
comparable to or exceeding that of conventional sparse MoE architectures while activating fewer experts on average. This
demonstrates that increased routing flexibility can yield improved compute efficiency without degrading model quality.

The results suggest that differentiable top-$k$ routing constitutes a viable and scalable alternative to hard expert
selection in MoE architectures. Beyond the specific LapSum-based relaxation introduced in this work, the proposed
framework enables further investigation into adaptive conditional computation, including finer-grained budget control,
interactions with scaling behavior, and extensions to larger-scale and multimodal models.

\section*{Limitations}

We evaluate our models on two English text corpora (OWT and C4) and English language downstream tasks. Evaluation of
additional languages and tasks would provide a broader picture of SoftMoE capabilities. Another avenue for exploration
not pursued in this work is applications of SoftMoE in multimodal models. Finally, our experiments are conducted on
a~1.63B parameter architecture. While substantial, it is smaller than the largest language models deployed in practice.
Our conclusions could, therefore, be strengthened by further validation at larger scales.

\section*{Impact Statement}

This paper presents work whose goal is to advance the field of machine learning. There are many potential societal
consequences of our work, none of which we feel must be specifically highlighted here.

\section*{Acknowledgments}

This work was funded in whole or in part by the National Science Centre, Poland, and The National Centre for Research
and Development, Poland, ARTIQ project: UMO-2021/01/2/ST6/00004 and ARTIQ/0004/2021, as well as by the "Excellence
initiative--research university" program for AGH University of Krakow.
Additionally, the work of Jacek Tabor and Łukasz Struski was supported by the National Science Centre, Poland, grants no.
2023/49/B/ST6/01137.

We gratefully acknowledge Polish high-performance computing infrastructure PLGrid (HPC Center: ACK Cyfronet AGH) for
providing computer facilities and support within computational grant no. PLG/2025/018349.

\bibliography{bibliography}

@inproceedings{berrada2018smoothlossfunctionsdeep,
  author       = {Leonard Berrada and Andrew Zisserman and M. Pawan Kumar},
  title        = {Smooth Loss Functions for Deep Top-k Classification},
  booktitle    = {6th International Conference on Learning Representations, {ICLR} 2018},
  publisher    = {OpenReview.net},
  year         = {2018},
}

@inproceedings{chen2019topkoffpolicy,
  author       = {Minmin Chen and Alex Beutel and Paul Covington and Sagar Jain and
                  Francois Belletti and Ed H. Chi},
  title        = {Top-K Off-Policy Correction for a {REINFORCE} Recommender System},
  booktitle    = {Proceedings of the Twelfth {ACM} International Conference on Web Search
                  and Data Mining, {WSDM} 2019},
  pages        = {456--464},
  publisher    = {{ACM}},
  year         = {2019},
}

@inproceedings{chen2023learningsparsetransformer,
  author       = {Xiang Chen and Hao Li and Mingqiang Li and Jinshan Pan},
  title        = {Learning {A} Sparse Transformer Network for Effective Image Deraining},
  booktitle    = {{IEEE/CVF} Conference on Computer Vision and Pattern Recognition,
                  {CVPR} 2023},
  pages        = {5896--5905},
  publisher    = {{IEEE}},
  year         = {2023},
}

@inproceedings{cortes2024cardinality,
  author       = {Corinna Cortes and Anqi Mao and Christopher Mohri and Mehryar Mohri and Yutao Zhong},
  title        = {Cardinality-Aware Set Prediction and Top-$k$ Classification},
  booktitle    = {Advances in Neural Information Processing Systems 37: Annual Conference
                  on Neural Information Processing Systems 2024, NeurIPS 2024},
  year         = {2024},
}

@inproceedings{cuturi2019differentiable,
  author       = {Marco Cuturi and Olivier Teboul and Jean{-}Philippe Vert},
  title        = {Differentiable Ranking and Sorting using Optimal Transport},
  booktitle    = {Advances in Neural Information Processing Systems 32: Annual Conference
                  on Neural Information Processing Systems 2019, NeurIPS 2019},
  pages        = {6858--6868},
  year         = {2019}
}

@inproceedings{DBLP:conf/aaai/BiskZLGC20,
  author    = {Yonatan Bisk and
               Rowan Zellers and
               Ronan Le Bras and
               Jianfeng Gao and
               Yejin Choi},
  title     = {{PIQA:} Reasoning about Physical Commonsense in Natural Language},
  booktitle = {The Thirty-Fourth {AAAI} Conference on Artificial Intelligence, {AAAI}
               2020},
  pages     = {7432--7439},
  publisher = {{AAAI} Press},
  year      = {2020}
}

@inproceedings{DBLP:conf/acl/ZellersHBFC19,
  author    = {Rowan Zellers and
               Ari Holtzman and
               Yonatan Bisk and
               Ali Farhadi and
               Yejin Choi},
  title     = {HellaSwag: Can a Machine Really Finish Your Sentence?},
  booktitle = {Proceedings of the 57th Annual Meeting of the Association for Computational Linguistics, {ACL} 2019, Volume 1: Long Papers},
  pages     = {4791--4800},
  publisher = {Association for Computational Linguistics},
  year      = {2019}
}

@inproceedings{DBLP:conf/iclr/LepikhinLXCFHKS21,
  author    = {Dmitry Lepikhin and
               HyoukJoong Lee and
               Yuanzhong Xu and
               Dehao Chen and
               Orhan Firat and
               Yanping Huang and
               Maxim Krikun and
               Noam Shazeer and
               Zhifeng Chen},
  title     = {GShard: Scaling Giant Models with Conditional Computation and Automatic Sharding},
  booktitle = {9th International Conference on Learning Representations, {ICLR} 2021},
  publisher = {OpenReview.net},
  year      = {2021}
}

@inproceedings{DBLP:conf/iclr/LiuPB23,
  author    = {Tianlin Liu and
               Joan Puigcerver and
               Mathieu Blondel},
  title     = {Sparsity-Constrained Optimal Transport},
  booktitle = {The Eleventh International Conference on Learning Representations, {ICLR} 2023},
  publisher = {OpenReview.net},
  year      = {2023}
}

@inproceedings{DBLP:conf/iclr/ZhaoC0C24,
  author    = {Xinyu Zhao and
               Xuxi Chen and
               Yu Cheng and
               Tianlong Chen},
  title     = {Sparse MoE with Language Guided Routing for Multilingual Machine Translation},
  booktitle = {The Twelfth International Conference on Learning Representations,
               {ICLR} 2024},
  publisher = {OpenReview.net},
  year      = {2024}
}

@inproceedings{DBLP:conf/icml/ClarkCGMPHDHCB022,
  author    = {Aidan Clark and
               Diego de Las Casas and
               Aurelia Guy and
               Arthur Mensch and
               Michela Paganini and
               Jordan Hoffmann and
               Bogdan Damoc and
               Blake A. Hechtman and
               Trevor Cai and
               Sebastian Borgeaud and
               George van den Driessche and
               Eliza Rutherford and
               Tom Hennigan and
               Matthew J. Johnson and
               Albin Cassirer and
               Chris Jones and
               Elena Buchatskaya and
               David Budden and
               Laurent Sifre and
               Simon Osindero and
               Oriol Vinyals and
               Marc'Aurelio Ranzato and
               Jack W. Rae and
               Erich Elsen and
               Koray Kavukcuoglu and
               Karen Simonyan},
  title     = {Unified Scaling Laws for Routed Language Models},
  booktitle = {Proceedings of the 39th International Conference on Machine Learning,
               {ICML} 2022},
  pages     = {4057--4086},
  publisher = {{PMLR}},
  year      = {2022}
}

@inproceedings{DBLP:conf/icml/LewisBDGZ21,
  author    = {Mike Lewis and
               Shruti Bhosale and
               Tim Dettmers and
               Naman Goyal and
               Luke Zettlemoyer},
  title     = {{BASE} Layers: Simplifying Training of Large, Sparse Models},
  booktitle = {Proceedings of the 38th International Conference on Machine Learning,
               {ICML} 2021},
  pages     = {6265--6274},
  publisher = {{PMLR}},
  year      = {2021}
}

@inproceedings{DBLP:conf/icml/LudziejewskiKAP24,
  author    = {Jan Ludziejewski and
               Jakub Krajewski and
               Kamil Adamczewski and
               Maciej Pi{\'{o}}ro and
               Michal Krutul and
               Szymon Antoniak and
               Kamil Ciebiera and
               Krystian Kr{\'{o}}l and
               Tomasz Odrzyg{\'{o}}zdz and
               Piotr Sankowski and
               Marek Cygan and
               Sebastian Jaszczur},
  title     = {Scaling Laws for Fine-Grained Mixture of Experts},
  booktitle = {Proceedings of the 41st International Conference on Machine Learning, {ICML} 2024},
  pages     = {33270--33288},
  publisher = {{PMLR}},
  year      = {2024}
}

@inproceedings{DBLP:conf/icml/StruskiBPT25,
  author       = {{\L}ukasz Struski and Michal B. Bednarczyk and Igor T. Podolak and Jacek Tabor},
  title        = {{L}ap{S}um - One Method to Differentiate Them All: Ranking, Sorting and Top-k Selection},
  booktitle    = {Proceedings of the 42nd International Conference on Machine Learning, {ICML} 2025},
  pages        = {56990--57007},
  publisher    = {{PMLR}},
  year         = {2025},
}

@inproceedings{DBLP:conf/nips/AntoniakKPKLCKO24,
  author    = {Szymon Antoniak and
               Michal Krutul and
               Maciej Pi{\'{o}}ro and
               Jakub Krajewski and
               Jan Ludziejewski and
               Kamil Ciebiera and
               Krystian Kr{\'{o}}l and
               Tomasz Odrzyg{\'{o}}zdz and
               Marek Cygan and
               Sebastian Jaszczur},
  title     = {Mixture of Tokens: Continuous {MoE} through Cross-Example Aggregation},
  booktitle = {Advances in Neural Information Processing Systems 37: Annual Conference
               on Neural Information Processing Systems 2024, NeurIPS 2024},
  year      = {2024}
}

@inproceedings{DBLP:conf/nips/MuennighoffRBST23,
  author    = {Niklas Muennighoff and
               Alexander M. Rush and
               Boaz Barak and
               Teven Le Scao and
               Nouamane Tazi and
               Aleksandra Piktus and
               Sampo Pyysalo and
               Thomas Wolf and
               Colin A. Raffel},
  title     = {Scaling Data-Constrained Language Models},
  booktitle = {Advances in Neural Information Processing Systems 36: Annual Conference on Neural Information Processing Systems 2023, NeurIPS 2023},
  year      = {2023}
}

@inproceedings{DBLP:conf/nips/RollerSSW21,
  author    = {Stephen Roller and
               Sainbayar Sukhbaatar and
               Arthur Szlam and
               Jason Weston},
  title     = {Hash Layers For Large Sparse Models},
  booktitle = {Advances in Neural Information Processing Systems 34: Annual Conference on Neural Information Processing Systems 2021, NeurIPS 2021},
  pages     = {17555--17566},
  year      = {2021}
}

@inproceedings{DBLP:conf/nips/ZhouLLDHZDCLL22,
  author    = {Yanqi Zhou and
               Tao Lei and
               Hanxiao Liu and
               Nan Du and
               Yanping Huang and
               Vincent Y. Zhao and
               Andrew M. Dai and
               Zhifeng Chen and
               Quoc V. Le and
               James Laudon},
  title     = {Mixture-of-Experts with Expert Choice Routing},
  booktitle = {Advances in Neural Information Processing Systems 35: Annual Conference
               on Neural Information Processing Systems 2022, NeurIPS 2022},
  year      = {2022}
}

@article{DBLP:journals/corr/abs-1909-08053,
  author     = {Mohammad Shoeybi and
                Mostofa Patwary and
                Raul Puri and
                Patrick LeGresley and
                Jared Casper and
                Bryan Catanzaro},
  title      = {Megatron-LM: Training Multi-Billion Parameter Language Models Using
                Model Parallelism},
  journal    = {arXiv preprint arXiv:1909.08053},
  year       = {2019},
}

@article{fedus2022switch,
  author  = {William Fedus and Barret Zoph and Noam Shazeer},
  title   = {Switch Transformers: Scaling to Trillion Parameter Models with Simple and Efficient Sparsity},
  journal = {Journal of Machine Learning Research},
  year    = {2022},
  volume  = {23},
  number  = {120},
  pages   = {1--39},
}

@article{DBLP:journals/corr/abs-2102-03315,
  author     = {Sumithra Bhakthavatsalam and
                Daniel Khashabi and
                Tushar Khot and
                Bhavana Dalvi Mishra and
                Kyle Richardson and
                Ashish Sabharwal and
                Carissa Schoenick and
                Oyvind Tafjord and
                Peter Clark},
  title      = {Think you have Solved Direct-Answer Question Answering? Try {ARC-DA},
                the Direct-Answer {AI2} Reasoning Challenge},
  journal    = {arXiv preprint arXiv:2102.03315},
  year       = {2021},
}

@inproceedings{riquelme2021scaling,
  author       = {Carlos Riquelme and
                  Joan Puigcerver and
                  Basil Mustafa and
                  Maxim Neumann and
                  Rodolphe Jenatton and
                  Andr{\'{e}} Susano Pinto and
                  Daniel Keysers and
                  Neil Houlsby},
  title        = {Scaling Vision with Sparse Mixture of Experts},
  booktitle    = {Advances in Neural Information Processing Systems 34: Annual Conference
                  on Neural Information Processing Systems 2021, NeurIPS 2021},
  pages        = {8583--8595},
  year         = {2021},
}

@article{DBLP:journals/corr/abs-2203-15556,
  author     = {Jordan Hoffmann and
                Sebastian Borgeaud and
                Arthur Mensch and
                Elena Buchatskaya and
                Trevor Cai and
                Eliza Rutherford and
                Diego de Las Casas and
                Lisa Anne Hendricks and
                Johannes Welbl and
                Aidan Clark and
                Tom Hennigan and
                Eric Noland and
                Katie Millican and
                George van den Driessche and
                Bogdan Damoc and
                Aurelia Guy and
                Simon Osindero and
                Karen Simonyan and
                Erich Elsen and
                Jack W. Rae and
                Oriol Vinyals and
                Laurent Sifre},
  title      = {Training Compute-Optimal Large Language Models},
  journal    = {arXiv preprint arXiv:2203.15556},
  year       = {2022},
}

@article{DBLP:journals/corr/abs-2404-05567,
  author     = {Bowen Pan and
                Yikang Shen and
                Haokun Liu and
                Mayank Mishra and
                Gaoyuan Zhang and
                Aude Oliva and
                Colin Raffel and
                Rameswar Panda},
  title      = {Dense Training, Sparse Inference: Rethinking Training of Mixture-of-Experts
                Language Models},
  journal    = {arXiv preprint arXiv:2404.05567},
  year       = {2024},
}

@article{DBLP:journals/corr/BengioBPP15,
  author     = {Emmanuel Bengio and
                Pierre{-}Luc Bacon and
                Joelle Pineau and
                Doina Precup},
  title      = {Conditional Computation in Neural Networks for faster models},
  journal    = {arXiv preprint arXiv:1511.06297},
  year       = {2015},
}

@inproceedings{DBLP:conf/iclr/ShazeerMMDLHD17,
  author       = {Noam Shazeer and
                  Azalia Mirhoseini and
                  Krzysztof Maziarz and
                  Andy Davis and
                  Quoc V. Le and
                  Geoffrey E. Hinton and
                  Jeff Dean},
  title        = {Outrageously Large Neural Networks: The Sparsely-Gated Mixture-of-Experts
                  Layer},
  booktitle    = {5th International Conference on Learning Representations, {ICLR} 2017},
  publisher    = {OpenReview.net},
  year         = {2017},
}

@article{DBLP:journals/jmlr/RaffelSRLNMZLL20,
  author  = {Colin Raffel and
             Noam Shazeer and
             Adam Roberts and
             Katherine Lee and
             Sharan Narang and
             Michael Matena and
             Yanqi Zhou and
             Wei Li and
             Peter J. Liu},
  title   = {Exploring the Limits of Transfer Learning with a Unified Text to-Text Transformer},
  journal = {Journal of Machine Learning Research},
  volume  = {21},
  number  = {140},
  pages   = {1--67},
  year    = {2020}
}

@article{DBLP:journals/neco/JacobsJNH91,
  author  = {Robert A. Jacobs and
             Michael I. Jordan and
             Steven J. Nowlan and
             Geoffrey E. Hinton},
  title   = {Adaptive Mixtures of Local Experts},
  journal = {Neural Comput.},
  volume  = {3},
  number  = {1},
  pages   = {79--87},
  year    = {1991}
}

@article{DBLP:journals/neco/JordanJ94,
  author  = {Michael I. Jordan and
             Robert A. Jacobs},
  title   = {Hierarchical Mixtures of Experts and the {EM} Algorithm},
  journal = {Neural Comput.},
  volume  = {6},
  number  = {2},
  pages   = {181--214},
  year    = {1994}
}

@inproceedings{garcin2022stochastic,
  author       = {Camille Garcin and
                  Maximilien Servajean and
                  Alexis Joly and
                  Joseph Salmon},
  title        = {Stochastic smoothing of the top-K calibrated hinge loss for deep imbalanced
                  classification},
  booktitle    = {Proceedings of the 39th International Conference on Machine Learning, {ICML} 2022},
  pages        = {7208--7222},
  publisher    = {{PMLR}},
  year         = {2022},
}

@misc{Gokaslan2019OpenWeb,
  title        = {{OpenWebText} Corpus},
  author       = {Gokaslan, Aaron and Cohen, Vanya},
  howpublished = {\url{http://Skylion007.github.io/OpenWebTextCorpus}},
  year         = {2019}
}

@article{hoefler2021sparsity,
  author   = {Torsten Hoefler and
                  Dan Alistarh and
                  Tal Ben{-}Nun and
                  Nikoli Dryden and
                  Alexandra Peste},
  title   = {Sparsity in Deep Learning: Pruning and growth for efficient inference and training in neural networks},
  journal = {Journal of Machine Learning Research},
  year    = {2021},
  volume  = {22},
  number  = {241},
  pages   = {1--124},
}

@article{kaplan2020scaling,
  author       = {Jared Kaplan and
                  Sam McCandlish and
                  Tom Henighan and
                  Tom B. Brown and
                  Benjamin Chess and
                  Rewon Child and
                  Scott Gray and
                  Alec Radford and
                  Jeffrey Wu and
                  Dario Amodei},
  title        = {Scaling Laws for Neural Language Models},
  journal      = {arXiv preprint arXiv:2001.08361},
  year         = {2020},
}

@inproceedings{lapin2016lossfunctionstopkerror,
  author       = {Maksim Lapin and
                  Matthias Hein and
                  Bernt Schiele},
  title        = {Loss Functions for Top-k Error: Analysis and Insights},
  booktitle    = {2016 {IEEE} Conference on Computer Vision and Pattern Recognition,
                  {CVPR} 2016},
  pages        = {1468--1477},
  publisher    = {{IEEE} Computer Society},
  year         = {2016},
}

@article{mao2024topkclassification,
  author       = {Anqi Mao and
                  Mehryar Mohri and
                  Yutao Zhong},
  title        = {Top-k Classification and Cardinality-Aware Prediction},
  journal      = {arXiv preprint arXiv:2403.19625},
  year         = {2024}
}

@article{masud2023multivariatesoftrank,
  author = {Shoaib Bin Masud and
                  Matthew Werenski and
                  James M. Murphy and
                  Shuchin Aeron},
  title  = {Multivariate Soft Rank via Entropy-Regularized Optimal Transport: Sample Efficiency and Generative Modeling},
  journal={Journal of Machine Learning Research},
  volume={24},
  number={160},
  pages={1--65},
  year={2023}
}

@inproceedings{puigcerver2024soft,
  author       = {Joan Puigcerver and
                  Carlos Riquelme Ruiz and
                  Basil Mustafa and
                  Neil Houlsby},
  title        = {From Sparse to Soft Mixtures of Experts},
  booktitle    = {The Twelfth International Conference on Learning Representations,
                  {ICLR} 2024},
  publisher    = {OpenReview.net},
  year         = {2024},
}

@misc{radford2019language,
  title   = {Language Models are Unsupervised Multitask Learners},
  author  = {Radford, Alec and Wu, Jeffrey and Child, Rewon and Luan, David and Amodei, Dario and Sutskever, Ilya},
  year    = {2019},
  howpublished = {OpenAI Blog},
  note    = {Accessed: 2024-11-15},
}

@inproceedings{xie2020differentiable,
  author       = {Yujia Xie and
                  Hanjun Dai and
                  Minshuo Chen and
                  Bo Dai and
                  Tuo Zhao and
                  Hongyuan Zha and
                  Wei Wei and
                  Tomas Pfister},
  title        = {Differentiable Top-k with Optimal Transport},
  booktitle    = {Advances in Neural Information Processing Systems 33: Annual Conference
                  on Neural Information Processing Systems 2020, NeurIPS 2020},
  year         = {2020},
}

@article{zhao2019explicitsparsetransformer,
  author       = {Guangxiang Zhao and
                  Junyang Lin and
                  Zhiyuan Zhang and
                  Xuancheng Ren and
                  Qi Su and
                  Xu Sun},
  title        = {Explicit Sparse Transformer: Concentrated Attention Through Explicit
                  Selection},
  journal      = {arXiv preprint arXiv:1912.11637},
  year         = {2019}
}

@article{zoph2022st,
  author  = {Barret Zoph and Irwan Bello and Sameer Kumar and Nan Du and Yanping Huang and Jeff Dean and Noam Shazeer and William Fedus},
  title   = {{ST-MoE}: Designing Stable and Transferable Sparse Expert Models}, 
  journal = {arXiv preprint arXiv:2202.08906},
  year    = {2022},
}

@inproceedings{tenney2019bert,
  title     = {{BERT} Rediscovers the Classical {NLP} Pipeline},
  author    = {Tenney, Ian  and Das, Dipanjan  and Pavlick, Ellie},
  booktitle = {Proceedings of the 57th Annual Meeting of the Association for Computational Linguistics, {ACL} 2019},
  year      = {2019},
  publisher = {Association for Computational Linguistics},
  pages     = {4593--4601},
}

@inproceedings{jawahar2019bert,
  title = {What Does {BERT} Learn about the Structure of Language?},
  author = {Jawahar, Ganesh  and Sagot, Beno{\^i}t  and Seddah, Djam{\'e}},
  booktitle = {Proceedings of the 57th Annual Meeting of the Association for Computational Linguistics, {ACL} 2019},
  year = {2019},
  publisher = {Association for Computational Linguistics},
  pages = {3651--3657}
}
\bibliographystyle{icml2026}

\newpage
\appendix
\onecolumn
\section{Learned expert allocation patterns} \label{app:additional_results}

\begin{figure}[thb]
    \centering
    \includegraphics[width=0.6\linewidth]{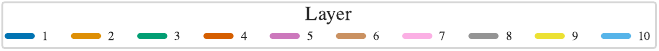}
    \includegraphics[width=1\linewidth]{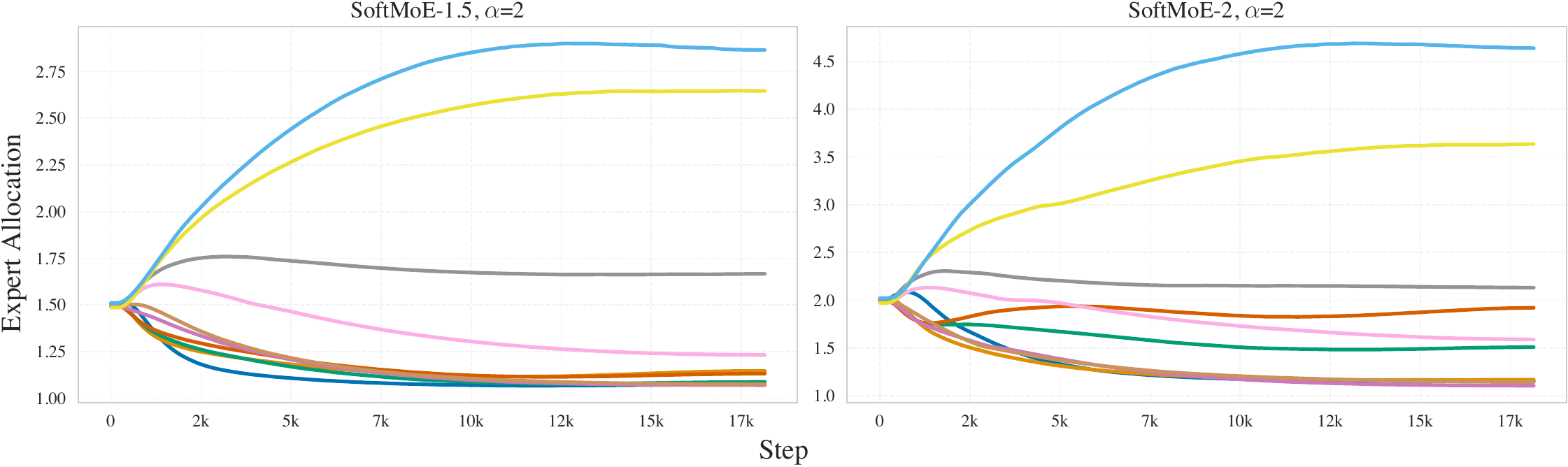}

    \caption{Evolution of the expert allocation parameters ($k_l$) during SoftMoE pretraining on OWT.}
\end{figure}

\begin{figure}[thb]
    \centering
    \includegraphics[width=0.6\linewidth]{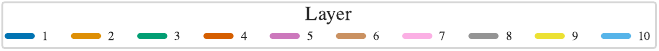}
    \includegraphics[width=1\linewidth]{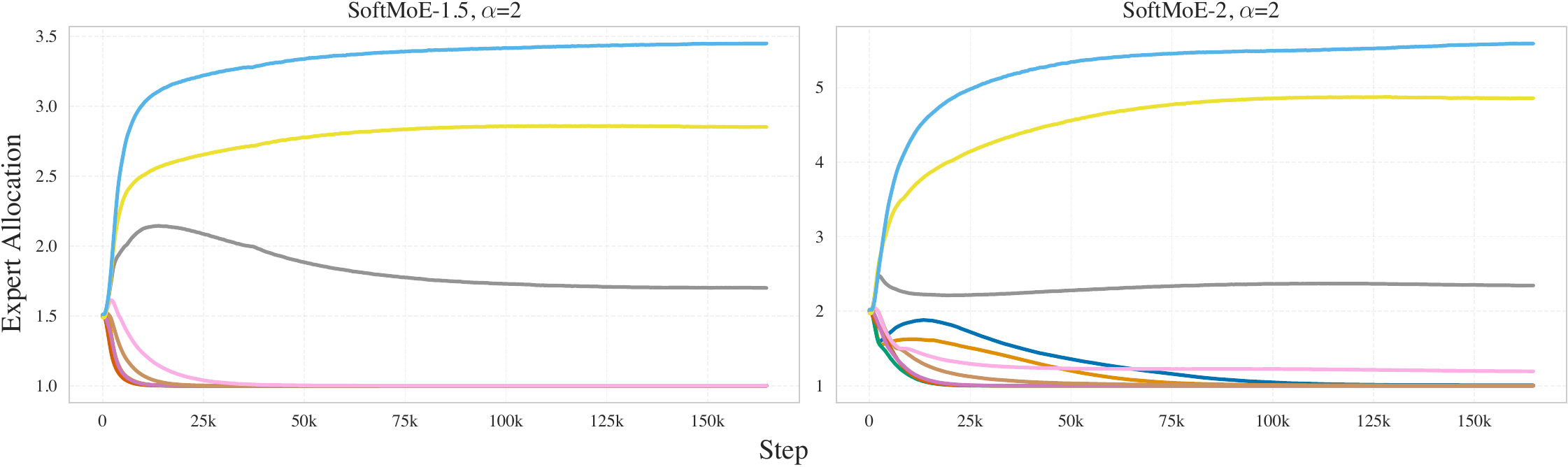}

    \caption{Evolution of the expert allocation parameters ($k_l$) during SoftMoE pretraining on C4.}
\end{figure}

\section{Training Hyperparameters} \label{app:train_param}

Configuration of models used in experiments is summarized in Table~\ref{tab:model-size-params}. All models were trained
with mixed floating point precision (bfloat16 and float32). Experiments on the OWT dataset were conducted with a global
batch size of 240, while those on the C4 dataset were conducted with a batch size of 600.  We used eight GPUs to train
the models on OWT and twenty GPUs to train on C4. In all configurations, we used a sequence length of 2048. The number
of training steps depends on the dataset, and was 18k and 164k iterations for OWT and C4, respectively. We used the Adam
optimizer with $\beta_1 = 0.9$ and $\beta_2 = 0.95$. The learning rate was annealed from $10^{-3}$ to $10^{-5}$ with a
cosine decay. Additionally, we used a warm-up phase in the initial $1\%$ of training. All models contained 32 experts.
An auxiliary loss function was used for intra-layer load-balancing. We also experimented with different jitter
approaches in the standard sparse MoE architecture. However, they led to degraded performance (Table~\ref{app:jitter}).
Consequently, jitter was disabled in our sparse MoE baselines to ensure a fairer comparison, making the reported
baselines stronger than configurations trained with jitter throughout. The LapSum-based $\SoftTopK$ implementation uses
a temperature hyper-parameter that was annealed during training with a cosine decay from 5 to 1.  We used RMSNorm in
normalization layers and SiLU activation function. We trained our models with 8-way data parallelism for OWT and 20-way
data parallelism for C4. Tensor parallelism and pipeline parallelism were not used. All experiments were computed on
NVIDIA GH200 CPU-GPU chips with 96GB of VRAM and 120GB RAM each. Computing resources used in training are summarized in
Table~\ref{tab:training_time_gpus}.

\begin{table*}[thb]
    \centering
    \caption{Hyperparameters used to define model size.}
    \label{tab:model-size-params}

    \begin{tabular}{lccccccc}
        \toprule
         Model & Total Params & Blocks & Att. Heads & Experts & $d_{attn}$ & $d_{fnn}$ \\
        \midrule
        All & 1.631B & 10 & 10 & 32 & 640 & 2560\\
        \bottomrule 
    \end{tabular}
\end{table*}

\begin{table*}[tbh]
\centering
  \caption{Computing resources used for model training.}
  \label{tab:training_time_gpus}
  \begin{tabular}{clcccc}
    \toprule
    Dataset & Model & Training Time (h) & Number of GPUs & GPUh\\
    \midrule
    \multirow{9}{*}{
          \rotatebox[origin=c]{90}{Open Web Text}
    }

    & Sparse MoE-1 & 3 & 8 & 23.7\\
    & Sparse MoE-2 & 2.2 & 8 & 17.7\\
    & Sparse MoE-3 & 2.3 & 8 & 18.5\\
    & Sparse MoE-4 & 2.6 & 8 & 20.6\\
    & SoftMoE*-1 & 3.2 & 8 & 25.7\\
    & SoftMoE*-1.5 & 2.5 & 8 & 20.0\\
    & SoftMoE-1.5 & 1.9 & 8 & 15.5\\
    & SoftMoE*-2 & 2.4 & 8 & 19.1\\
    & SoftMoE-2 & 2.4 & 8 & 19.4\\

    \midrule
    \multirow{9}{*}{
          \rotatebox[origin=c]{90}{C4}
    }

    & Sparse MoE-1 & 18.4 & 20 & 367.1\\
    & Sparse MoE-2 & 19.5 & 20 & 390.4\\
    & Sparse MoE-3 & 20.7 & 20 & 413.7\\
    & Sparse MoE-4 & 23.6 & 20 & 472.4\\
    & SoftMoE*-1 & 25.4 & 20 & 507.3\\
    & SoftMoE*-1.5 & 21.2 & 20 & 424.3\\
    & SoftMoE-1.5 & 23.8 & 20 & 475.0\\
    & SoftMoE*-2 & 24.7 & 20 & 493.5\\
    & SoftMoE-2 & 25.4 & 20 & 508.1\\
    \bottomrule
  \end{tabular}
\end{table*}

\section{Additional Results}

\begin{table*}[tbh]
\centering
\caption{Comparison of the number of active experts, wall-clock training time, iteration time, throughput, and GPU
utilization. The throughput and training times in these experiments are influenced by the computational cluster’s load,
as reflected in the GPU utilization.}
\begin{tabular}{clccccccc}
\toprule
Dataset & Model & Train-AE & Time(h) & GPUh & Avg Iter Time(ms) & Tokens/s/gpu & GPU Util.(\%) \\
\midrule

\multirow{3}{*}{
      \rotatebox[origin=c]{90}{OWT}
}
& SparseMoE-2   & 2.00 & 2.2 & 17.7 & 438.96 & 142836 & 89.72 \\
& SoftMoE*-1.5 & 1.53 & 2.5 & 20.0 & 495.44 & 126554 & 82.75 \\
& SoftMoE-1.5  & 1.63 & 1.9 & 15.5 & 384.93 & 162883 & 94.02 \\

\bottomrule
\end{tabular}
\end{table*}

\vfill

\begin{table*}[tbh]
\centering
\caption{Relative variance in expert load (number of tokens per expert) in the SparseMoE and SoftMoE configurations.
Lower values indicate a more balanced distribution of experts. In our experiments, SoftMoE shows no discernible impact on load
balancing.}
\begin{tabular}{clccc}
\toprule
Dataset & Model & $k$ & $\alpha$ & Load variance (relative) \\
\midrule

\multirow{9}{*}{
      \rotatebox[origin=c]{90}{OWT}
}
& SparseMoE & 1   & -- & 0.10 \\
& SparseMoE & 2   & -- & 0.05 \\
& SoftMoE*  & 1.5 & 2  & 0.05 \\
& SoftMoE   & 1.5 & 2  & 0.05 \\
& SparseMoE & 3   & -- & 0.04 \\
& SparseMoE & 4   & -- & 0.04 \\
& SoftMoE*  & 1   & 4  & 0.03 \\
& SoftMoE*  & 2   & 2  & 0.07 \\
& SoftMoE   & 2   & 2  & 0.06 \\

\midrule

\multirow{9}{*}{
      \rotatebox[origin=c]{90}{C4}
}
& SparseMoE & 1   & -- & 0.04 \\
& SparseMoE & 2   & -- & 0.04 \\
& SoftMoE*  & 1.5 & 2  & 0.04 \\
& SparseMoE & 3   & -- & 0.03 \\
& SoftMoE   & 1.5 & 2  & 0.04 \\
& SparseMoE & 4   & -- & 0.03 \\
& SoftMoE*  & 1   & 4  & 0.02 \\
& SoftMoE*  & 2   & 2  & 0.03 \\
& SoftMoE   & 2   & 2  & 0.05 \\

\bottomrule
\end{tabular}
\end{table*}

\begin{table*}[tbh]
\centering
\caption{Effect of truncation threshold $\tau$ on the downstream performance of SoftMoE* on the OWT dataset. The effective
threshold is equal to $\tau \cdot k / E$, where $E$ is the number of experts and $k$ is the per-layer expert
allocation.}
\begin{tabular}{clcccc}
\toprule
Configuration & $\tau$ & AE & PIQA & Hella & ARC-E \\
\midrule

\multirow{3}{*}{$k=1,\alpha=4$}
& 1.5 & 2.47 & 62.35 & 32.78 & 36.51\\
& 1.3 & 3.10 & 64.04 & 32.97 & 34.39\\
& 1.1 & 3.73 & 63.93 & 33.68 & 36.51\\

\midrule

\multirow{3}{*}{$k=1.5,\alpha=2$}
& 1.8 & 1.73 & 62.63 & 32.13 & 35.10\\
& 1.5 & 2.01 & 62.02 & 31.81 & 37.04\\
& 1.3 & 2.35 & 63.87 & 32.16 & 37.04\\

\midrule

\multirow{4}{*}{$k=2,\alpha=2$}
& 1.8 & 1.81 & 62.57 & 31.34 & 35.27\\
& 1.5 & 2.28 & 62.62 & 31.88 & 35.10\\
& 1.3 & 2.70 & 62.57 & 32.54 & 35.10\\
& 1.2 & 3.12 & 63.82 & 32.50 & 37.21\\

\bottomrule
\end{tabular}
\end{table*}

\clearpage

\begin{table*}[tbh]
\centering
\caption{Ablation of truncation effects during inference-time. Results are reported for SoftMoE-2 trained with fixed
$\tau=1.2$ and evaluated across different truncation levels in inference. The effective threshold is equal to $\tau \cdot k_l /
E$, where $E$ is the number of experts and $k_l$ is the learned expert allocation for layer $l$.}
\begin{tabular}{clcccc}
\toprule
Dataset & $\tau$ & AE & PIQA & Hella & ARC-E \\
\midrule

\multirow{4}{*}{OWT}
& 2.4 & 1.71 & 62.94 & 32.33 & 36.68\\
& 1.8 & 2.21 & 63.49 & 32.82 & 37.56\\
& 1.2 & 3.50 & 63.55 & 32.74 & 38.27\\
& 0.6 & 4.52 & 63.65 & 32.69 & 37.21\\

\midrule

\multirow{4}{*}{C4}
& 2.4 & 1.62 & 71.00 & 46.98 & 40.03\\
& 1.8 & 1.68 & 71.65 & 47.56 & 41.44\\
& 1.2 & 3.75 & 72.03 & 47.48 & 42.15\\
& 0.6 & 4.69 & 71.92 & 47.47 & 41.62\\

\bottomrule
\end{tabular}
\end{table*}

\begin{table*}[tbh]
\centering
\caption{Effects of the routing jitter in sparse MoE baselines. In most cases, jitter degrades baseline performance.
}
\label{app:jitter}
\begin{tabular}{clccccc}
\toprule
Jitter & Model & PIQA & Hella & ARC-E & Loss \\
\midrule

No & Top-1 & 60.77 & 28.79 & 36.86 & 3.06 \\
No & Top-2 & 61.20 & 30.02 & 36.50 & 2.99 \\
No & Top-3 & 61.86 & 30.02 & 34.39 & 2.97 \\

\midrule

0.05 & Top-1 & 60.44 & 29.21 & 34.92 & 3.05 \\
0.05 & Top-2 & 60.72 & 29.08 & 33.86 & 3.00 \\
0.05 & Top-3 & 61.26 & 30.12 & 35.27 & 2.97 \\

\bottomrule
\end{tabular}
\end{table*}

\section{Contributions}
Mikołaj was responsible for method development, implementation of both the proof-of-concept and final versions of the
SoftMoE model, execution of experiments, performance evaluation, development of the research infrastructure and datasets
preparation. Marcin proposed the initial research concept, contributed to algorithm development, assisted in analysing
results and suggesting experiments, and coordinated the project. Łukasz was responsible for implementing the soft
top-$k$ algorithm, supported its understanding and adaptation in the system, and made contributions to the development
of the method. Jacek addressed theoretical aspects of the work, proposed unconstrained parametrization of the budget
allocator, and contributed to research direction and algorithm development. All authors contributed to the manuscript.


\end{document}